\documentclass[APA,STIX2COL]{WileyNJD-v2}

\usepackage{balance}
\usepackage{graphicx}
\usepackage{caption}
\usepackage{subcaption}
\makeatletter
\def\NAT@aysep{,}
\makeatother

\articletype{ORIGINAL ARTICLE}%

\received{Aug 2019}
\revised{xxx}
\accepted{xxx}

\begin{document}

\title{Investigation of wind pressures on tall building under interference effects using machine learning techniques}

\author[1]{Gang Hu}

\author[2,3]{Lingbo Liu}

\author[3]{Dacheng Tao}

\author[4]{Jie Song}

\author[1]{K.C.S. Kwok*}

\authormark{Hu \textsc{et al}}

\address[1]{\orgdiv{Centre for Wind, Waves and Water, School of Civil Engineering}, \orgname{The University of Sydney}, \orgaddress{\state{NSW 2006}, \country{Australia}}}

\address[2]{\orgdiv{School of Data and Computer Science}, \orgname{Sun Yat-sen University}, \orgaddress{\state{Guangzhou 510006}, \country{China}}}

\address[3]{\orgdiv{School of Computer Science}, \orgname{The University of Sydney}, \orgaddress{\state{NSW 2006}, \country{Australia}}}

\address[4]{\orgdiv{School of Civil Engineering}, \orgname{Wuhan University}, \orgaddress{\state{Wuhan 430072}, \country{China}}}

\corres{K.C.S. Kwok,  \email{kenny.kwok@sydney.edu.au}}

\abstract[Abstract]{Interference effects of tall buildings have attracted numerous studies due to the boom of clusters of tall buildings in megacities. To fully understand the interference effects of buildings, it often requires a substantial amount of wind tunnel tests. Limited wind tunnel tests that only cover part of interference scenarios are unable to fully reveal the interference effects. This study used machine learning techniques to resolve the conflicting requirement between limited wind tunnel tests that produce unreliable results and a completed investigation of the interference effects that is costly and time-consuming. Four machine learning models including decision tree, random forest, XGBoost, generative adversarial networks (GANs), were trained based on 30\% of a dataset to predict both mean and fluctuating pressure coefficients on the principal building. The GANs model exhibited the best performance in predicting these pressure coefficients. A number of GANs models were then trained based on different portions of the dataset ranging from 10\% to 90\%. It was found that the GANs model based on 30\% of the dataset is capable of predicting both mean and fluctuating pressure coefficients under unseen interference conditions accurately. By using this GANs model, 70\% of the wind tunnel test cases can be saved, largely alleviating the cost of this kind of wind tunnel testing study.}


\jnlcitation{\cname{%
\author{Gang Hu}, 
\author{Lingbo Liu}, 
\author{Dacheng Tao}, 
\author{Jie Song},
\author{K.C.S. Kwok}}, 
\ctitle{Investigation of wind pressures on tall building under interference effects using machine learning techniques}. \cjournal{xxxxxx}, \cvol{xxxxxx}.}

\maketitle

\section{Introduction}\label{sec1}

Groups of tall buildings are quite common in central business districts of megacities around the world. When wind passes through these buildings, their interference between each other may dramatically reduce wind loads acting on the buildings or conversely significantly amplify the loads. The amplification of overall wind loads may lead to severe vibrations of buildings, which causes serious occupant discomfort ~\citep{Kwok2009}. Furthermore, the amplification of local cladding wind pressures may induce, sometimes extensive, cladding damage and hence cause non-negligible economic loss, even threaten human lives~\citep{Kareem1986}.  To alleviate or eliminate the adverse interference effects of tall buildings, numerous studies have been devoted to exploring interference effects of tall buildings and reveal the underlying mechanisms. ~\cite{Khanduri1998} made a comprehensive review of studies of wind-induced interference effects on tall buildings. Four main parameters affecting interference between buildings were extracted, including type of upstream terrain, shape and size of buildings, incident wind direction, and building arrangement and spacing. The main research methodology in the past is based on wind tunnel tests.

Aeroelastic testing in the wind tunnel has been proven effective to study wind-structure interactions of tall buildings under interference from neighboring buildings. \cite{Bailey1985} studied interference excitation of two identical tall buildings by using aeroelastic tests in the wind tunnel. It was found that the dynamic loads on the upstream building may increase by a factor of up to 4.4 at a critical location of ($X, Y$) = ($-1.5b, 1.22b$), while that of the downstream building may increase by a factor of up to 3.2. \cite{Tang2004} utilized an aeroelastic 3DOF test rig to investigate the interference effects on wind-induced vibrations of tall buildings in three directions including alongwind, crosswind, and torsional. The dynamic torsional responses were also increased due to the wake of an upstream interfering building. Recently, \cite{Lo2016} used aeroelastic testing to identify another critical location of ($X, Y$) = ($-2b, 0$) for interference between two identical high-rise buildings, where the building has very strong crosswind vibration. All these studies only focused on the generic building shapes like square prism or rectangular prism buildings. \cite{Lo2017} studied interference effects of tall buildings with aerodynamic modification treatments. The principal building was modified by aerodynamic treatments while the neighboring building was a square prism model. They found that multiple modification treatment is efficient in reducing wind forces for all interference locations but in certain critical conditions the treatment may increase wind-induced vibrations. 

Pressure measurement and high frequency force balance (HFFB) tests were also used to measure wind forces acting on tall buildings subject to interference effects from adjacent buildings. \cite{Kim2011} examined local peak pressure coefficients of a building under interference of another building via pressure measurement tests in the wind tunnel. It was found that the highest peak suctions on the principal building increased with increasing height ratios of the interfering building and the staggered configuration produced larger peak suction than the tandem configuration. They also systematically studied overall wind loads on a building under interference of another building with identical cross-section but different height \citep{Kim2015}. The position and height of the interfering building were found to play an important role in generating base moments of the principal building. \cite{Mara2014} utilized HFFB tests to assess interference effects of a tall building on an identical building. The interference factors (IF) of wind-induced loads and responses were studied and the underlying mechanisms were discussed. \cite{Yan2016} tested a model of a real building with aerodynamic modifications under interference of an identical one via pressure measurements. The results show that critical tandem and staggered arrangements dramatically increase dynamic wind loads and responses, and the peak negative pressure coefficients are 1.3 times that without interference. 

To fully understand the crosswind excitation aerodynamic mechanisms of interference effects between two identical buildings, both pressure measurement tests and particle image velocimetry tests were conducted by \cite{Zu2018}. Synchronization of five quasi-periodic aerodynamic phenomena was observed and considered to be responsible for the enhancement of fluctuating crosswind forces of the principal building when the interfering building is placed at a longitudinal spacing of 5 building breaths ($D$) and a lateral spacing of 2.5$D$. The cross-sections of the interfering building and principal building in these studies are identical. \cite{Hui2017} studied interference effects on torsional moments between two buildings with identical height but different cross-section widths. The cross-section width ratio of the two buildings was 3. The mean torsion of a building under interference effect was up to 3 times of an isolated building, while the fluctuating torsion was 1.8 times that of an isolated building. \cite{Yu2018} also tested interference effects between two tall buildings with different widths of cross-section but identical height by using pressure measurement tests. They found that when the cross-section width ratio of the interfering building to the principal building equals to 0.4 to 0.6, the maximum alongwind enveloped interference factors reached 2.3 and 2.1, while the maximum crosswind enveloped interference factors were 3.0 and 2.1 due to vortex-induced resonance. 

Evidently, wind tunnel test techniques have advanced understanding of the interference effects of buildings. To fully under the interference effects of buildings, a huge amount of wind tunnel tests is required. Take an aerodynamic database pertaining to wind-induced interference effects of tall buildings constructed by Tokyo Polytechnic University (\url{http://wind.arch.t-kougei.ac.jp/system/eng/contents/code/tpu}) as an example, 37 locations of the interfering building and 72 wind attack angles for each location, representing in total 2664 cases, were tested. However, wind tunnel tests are inherently not only costly but also time-consuming. This kind of systematical parametric tests required substantial human resources and financial supports, which are not always affordable. Naturally, test cases in the wind tunnel studies are usually carefully selected and only able to cover part of representative scenarios. More specifically, wind tunnel tests are only able to test a series of discrete and very limited locations of the interfering building related to the principal building \citep{Flaga2018, Hui2012, lam2011, Song2016, Thepmongkorn2002, Yan2016}. Although linear interpolation methods are often adopted to generate an interference factor map, the accuracy of the interpolated value for untested scenarios is questionable since wind loading acting on the principal building is not as simple as a linear relationship with the relative position of the interfering building. Therefore, an interference map generated in this way is rather coarse and may not be accurate enough. What's more challenging is to obtain cladding pressures on the principal building of untested scenarios which represents localized pressure loading at discrete locations on a building instead of wind load effects acting on and wind-induced response of the principal building which represents an integrated wind effect and a structural response to this integrated wind effect. Likewise, the regular interpolation method is unsuited for such kind of task. Thus, it is beneficial to find a solution to systematically investigate the interference effects of buildings based on limited wind tunnel tests.

Machine learning (ML) techniques have been proven very effective to undertake regression and classification tasks in very complex problems. These techniques have exhibited huge potential of application in various engineering fields, such as civil engineering \citep{Adeli2001, Cha2017, Chou2015, Li2019, Maeda2018, Reich1996, Reich1997}, structural health monitoring \citep{Figueiredo2011, Ni2005, Santos2016, Wang2019, Worden2007}, construction materials \citep{Cheng2012, Chou2014, Sonebi2016}, and transportation engineering \citep{Bin2006, Liu2019, Liu2018a, Liu2018b, Lv2015}. However, the application of ML techniques in wind engineering is still very limited \citep{Bre2018, Fu2006, Fu2007, Huang2017}. \cite{Li2018} innovatively employed a ML approach to model vortex-induced vibrations (VIV) of a suspension bridge. A decision tree algorithm was adopted to train the VIV mode classification model and a support vector regression (SVR) algorithm was used to model the VIV response of the bridge deck. The classification and regression models can accurately identify and predict the VIV response for various modes of the bridge. A pioneering study by \cite{Jin2018} built a deep learning model for predicting the velocity field around a circular cylinder via fusion convolutional neural networks based on measured pressure field on the cylinder. The model was proven accurate when compared with CFD results and furthermore it successfully learned the underlying flow regimes of the cylinder. \cite{Oh2019} interestingly employed convolutional neural network (CNN) to estimate wind-induced responses of tall buildings. Top-level wind-induced displacement in the time and frequency domain, and wind data in the frequency domain were set as input of the CNN model. The maximum and minimum strains of columns were set as output of the model. The performance of the model was verified through strains not used in the CNN training. 

As mentioned above, to fully understand the interference effects of buildings at present, it requires a huge amount of wind tunnel tests, which is not always feasible due to high cost of wind tunnel tests. However, limited wind tunnel tests, which only cover part of representative scenarios, are unable to fully understand the interference effects. ML provides a promising solution to address this issue. This study aims to fully reveal the interference effects on buildings under various interference conditions including untested conditions based on limited wind tunnel test dataset by using ML techniques. The ML model will be trained based on wind pressures on buildings obtained from limited wind tunnel tests. A number of ML algorithms, including decision trees, random forest, XGBoost, and generative adversarial networks (GANs), will be tested. Their performances will be compared in order to select the best ML model in this study. The best model could be used to predict wind pressure on buildings under any interference conditions including untested conditions. As a result, the interference effects of buildings could be exposed fully. Furthermore, the successful accomplishment of this study can guide the application of ML to similar issues in wind engineering and even other relevant fields.

\section{Wind pressure database for high-rise building with adjacent building}\label{sec2}

\begin{figure*}[h!]
  \centering
  \includegraphics[width=\linewidth]{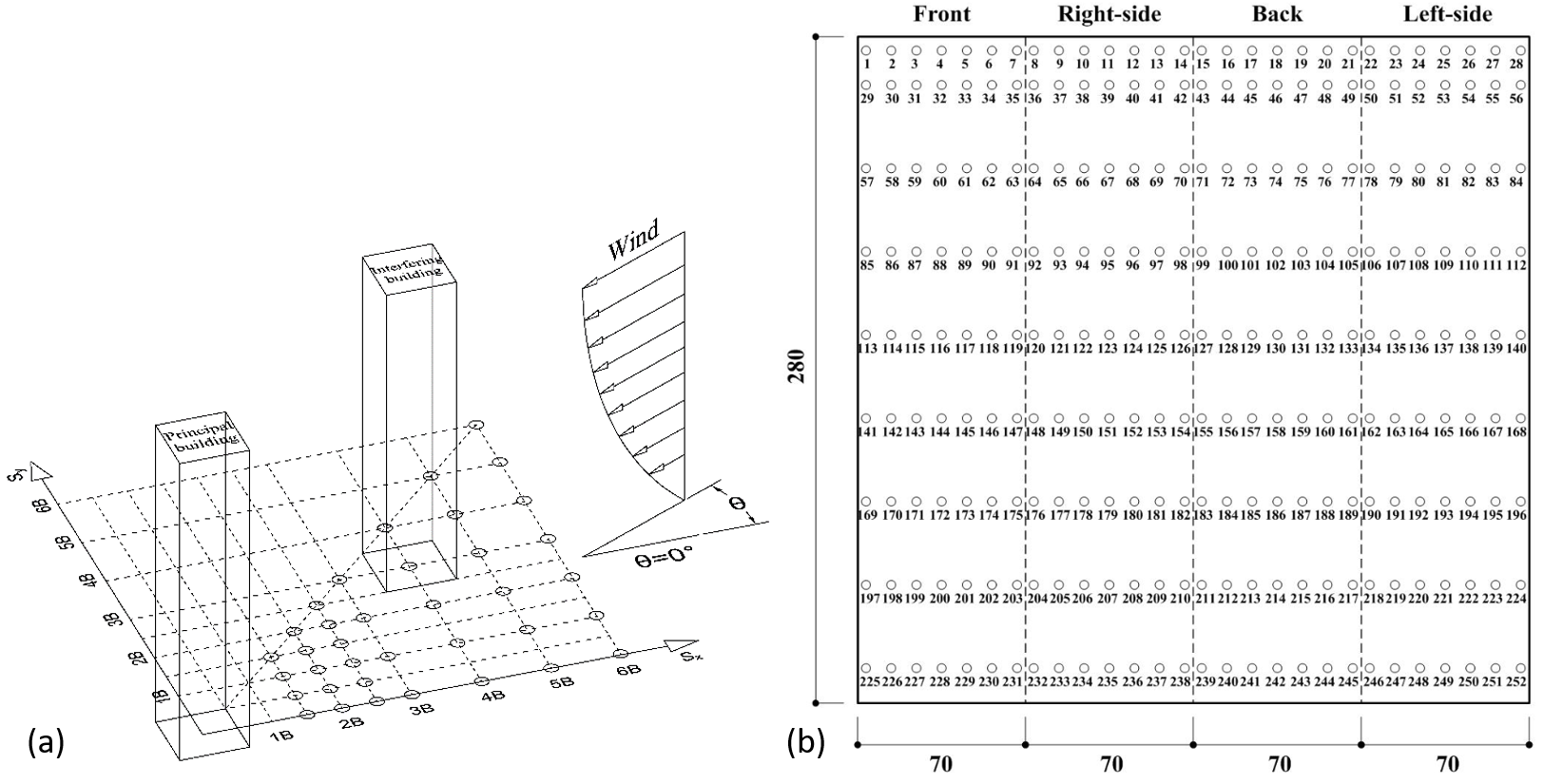}
  \caption{~(a) Relative locations of interfering building to principal building as tested and the test results stored in the database; (b) pressure taps on the principal building (\url{http://wind.arch.t-kougei.ac.jp/system/eng/contents/code/tpu}).}
  \label{fig1}
\end{figure*}

\begin{figure}
     \centering
     \begin{subfigure}{1\linewidth}
         \centering
         \includegraphics[width=\textwidth]{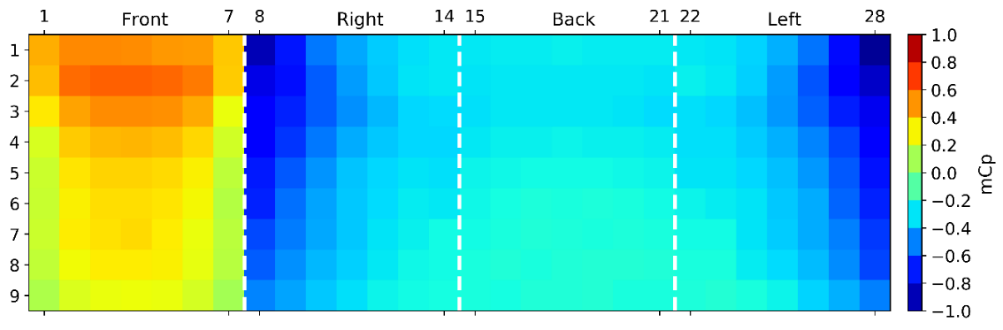}
         \caption{Mean pressure coefficient (mCp)}
         \label{fig2a}
     \end{subfigure}
     \hfill
     \begin{subfigure}{1\linewidth}
         \centering
         \includegraphics[width=\textwidth]{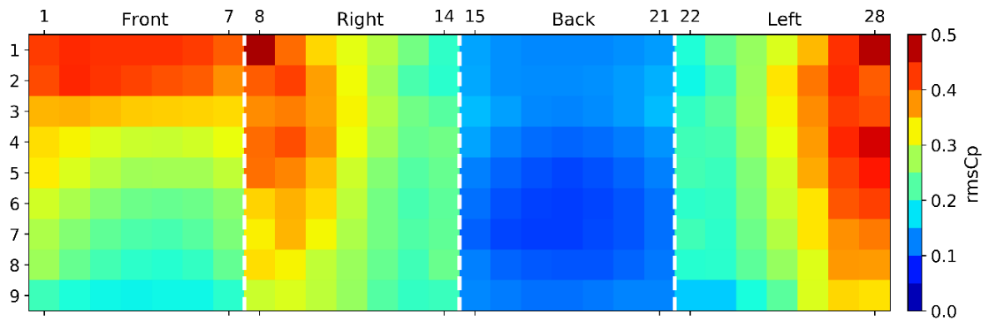}
         \caption{Fluctuating pressure coefficient (rmsCp)}
         \label{fig2b}
     \end{subfigure}
     \hfill
     \caption{~Mean and fluctuating pressure coefficients on principal building with interfering building located at $S_x = 6.0B, S_y = 0B$ and under zero wind attack angle.}
     \label{fig2}
\end{figure}

In this study, an aerodynamic database pertaining to wind-induced interference effects of tall buildings constructed by Tokyo Polytechnic University was utilized to demonstrate the feasibility of exploiting ML to predict wind pressures on buildings and reveal the interference effects under any interference conditions including untested conditions. This database provides wind tunnel testing data of wind pressures on surface of a tall building under interference of another tall building. In the wind tunnel tests, two tall building  models: a model with pressure taps distributed on its surface, referred to as the principal building of 70 m breadth, 70 m depth, and 280 m height at 1/400 length scale, and a dummy model, referred to as the interfering building, as shown in Figure~\ref{fig1}(a). 252 pressure taps were distributed on the principal building surfaces with 9 rows and 7 columns on each face, as shown in Figure~\ref{fig1}(b). The interfering building was located at 37 different upstream configurations with respect to the principal building, as shown in Figure~\ref{fig1}(a). 72 wind directions were considered, from 0$^\circ$ to 355$^\circ$ in 5$^\circ$ intervals, for each upstream configuration. Therefore, totally 2664 cases were tested in the wind tunnel. The database provides time series data of wind pressure coefficients at each test point on the building surface. More information can be found in the website (\url{http://wind.arch.t-kougei.ac.jp/system/eng/contents/code/tpu}) and relevant publications of the research group \citep{Kim2011,Kim2015}. In this study, the dataset of the interfering building with height identical to the principal building was used. Mean and fluctuating pressures on the principle building with the interfering building located at $S_x = 6.0B, S_y = 0$ and under zero wind attack angle are shown in Figure~\ref{fig2} as an example. 

The aforementioned 2664 cases with 37 different upstream configurations and 72 wind directions represent a comprehensive dataset to investigate the interference effects. To demonstrate the feasibility of using ML to predict wind pressures on buildings based on limited wind tunnel test data, only a portion of the 2664 cases will be used in training the ML model. First, 6 cases were randomly selected and set aside from the 2664 cases. These 6 cases will be used to validate machine learning predictions in the end. Second, 30\% of the remaining 2558 cases will be used to evaluate the capability of ML in predicting wind pressures on the principal building. Next, various portion of the entire database will be tested to demonstrate the effects of dataset on the performance of ML. 

\section{Machine learning model training}\label{sec3}
In this study, four machine learning (ML) algorithms, including decision tree regressor (DTR), random forest (RF), XGBoost (XGB), and generative adversarial networks (GANs), were employed to train models for predicting mean and fluctuating pressure coefficients on the principal building subjected to interference from an identical building nearby. 

A number of methods have been developed to evaluate the performance of ML models, such as re-substitution, hold-out, cross-validation (CV), and bootstrap \citep{Reich1999}. 10-fold CV method has been proven effective in evaluating the performance of the ML models \citep{Hu2019} and hence was used in this study. Optimizing the hyperparameters of ML algorithms is indispensable during the model training process. Considering the aforementioned 10-fold CV method and this optimization procedure, a 3-stage evaluation process called training-testing-validation (TTV) proposed by \cite{Reich1999} was adopted in this study. This TTV process comprises of three steps. The first step divides the dataset into training dataset and testing dataset. In this study, 80\% of the dataset were used to train the ML model and the rest were used to test the model. The second step selects the best ML algorithms and tunes hyperparameters using the 10-fold CV method. The last step builds a model based on the entire training dataset by using the best ML algorithms and best hyperparameters. This model is then validated by the testing dataset, i.e. 20\% of the whole dataset. 

\subsection{Decision tree}
The decision tree method trains a supervised ML model whereby the local region is identified in a sequence of recursive splits in a smaller number of steps \citep{Alpaydin2014}. A decision tree consists of root node, internal decision nodes and terminal leaves. The classification and regression tree (CART) algorithm, one of the algorithms of implementing decision trees \cite{Breiman1984}, was utilized to construct decision trees in the present study. The Gini index was used as its impurity function \citep{Loh2011}.

\begin{figure}[h!]
  \centering
  \includegraphics[width=\linewidth]{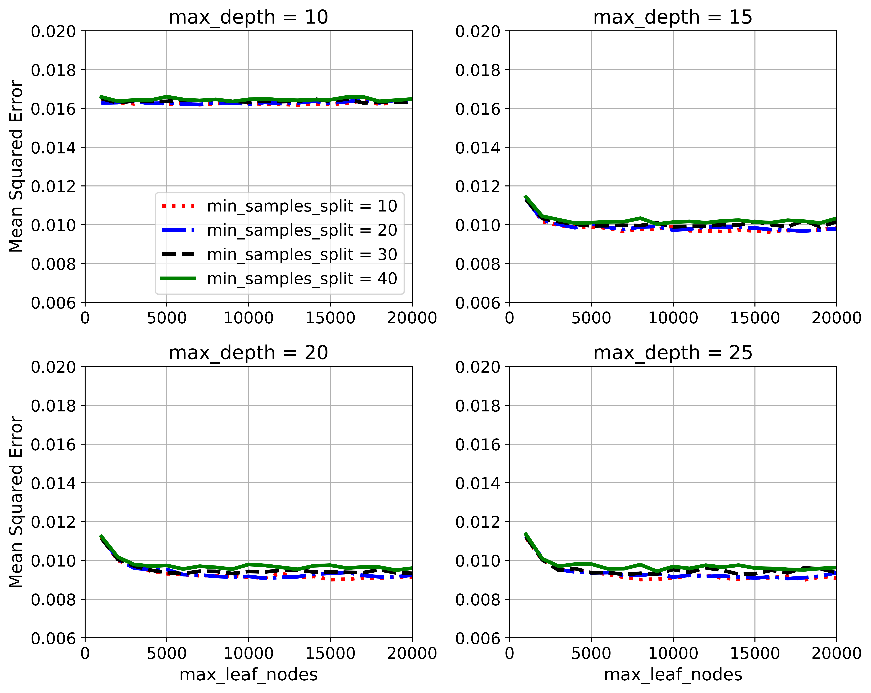}
  \caption{~Variations of 10-fold MSE with hyperparameters of DTR for dataset of mean pressure coefficients.}
  \label{fig3}
\end{figure}

During the model training process in the present study, effects of the hyperparameters, including the maximum depth of the tree (max\_depth), the maximum leaf nodes, and the minimum number of samples required to be at a leaf node (min\_samples\_leaf), on the performance of the model for the dataset of mean pressure coefficients were evaluated as shown in Figure~\ref{fig3}. It can be seen that when the maximum depth and the maximum leaf nodes reaches 20 and 20000 respectively, and the min\_samples\_leaf equals 20, the 10-fold mean squared error (MSE) reaches the minimum. Therefore, 20, 20000, and 20 were chosen for the above three parameters respectively in the decision tree regressor (DTR) to predict the mean pressure coefficients in this study. The similar hyperparameter optimization process was performed on the dataset of fluctuating pressure coefficients. It was found 25, 25000, and 15 are the optimal values for the three parameters respectively in predicting fluctuating pressure coefficients.

\subsection{Random forest}
Random forest (RF) is one of the most widely used ensemble method which aggregating multiple weak learners, e.g. decision trees, to overcome weaknesses of a single learner such as sensitivity to training data and unstableness \citep{Zhou2012}.  RF selects $n$ features among the total $m$ features for the split in each node. The number $n$ is set the same for all prediction trees, and it is recommended to be $1/3m$ \citep{Breiman2001}. The remainder of the algorithm is similar to the CART algorithm \citep{Breiman1984}.
\begin{figure}[h!]
  \centering
  \includegraphics[width=\linewidth]{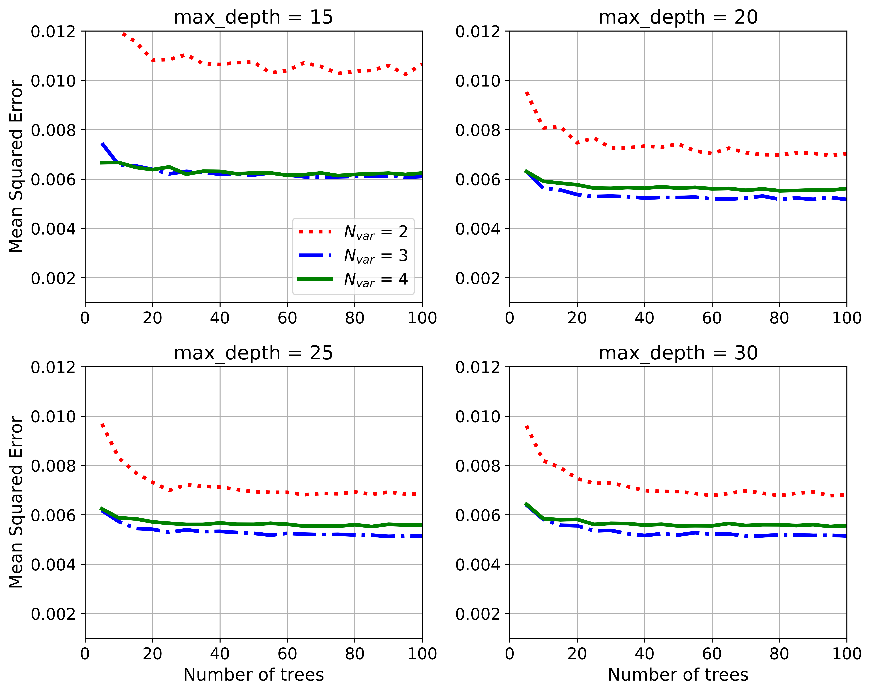}
  \caption{~Variations of 10-fold MSE with hyperparameters of RF for dataset of mean pressure coefficients.}
  \label{fig4}
\end{figure}

The optimization of hyperparameters of RF for mean pressure coefficients is shown in Figure~\ref{fig4}. It can be seen that the 10-fold MSE varies with number of grown trees, number of selected features, and the maximum depth of trees. MSE drops a lot with increasing maximum depth of trees up to 20 and a further increase in the maximum depth of trees induces negligible reduction in MSE. Therefore, 25 was selected to be the optimal maximum depth of trees. The number of selected features of 3 leads to the minimum MSE compared to 2 and 4 selected features. Furthermore, MSE is generally stable as the number of trees reaches 100. Therefore, the number of grown trees, the number of selected features, and the maximum depth were set as 100, 3, and 25 respectively for the dataset of mean pressure coefficients. Similarly, it was found that 150, 3, and 25 are the optimal values for the three parameters respectively for the dataset of fluctuating pressure coefficients.

\subsection{XGBoost}
Similar to random forest, XGBoost (XGB), short for eXtreme Gradient Boosting, is also one of the most popular ensemble methods and it is developed by \cite{Chen2016} very recently. According to XGBoost official documentation website (\url{https://xgboost.readthedocs.io/en/latest/}), XGB is an optimized distributed gradient boosting algorithm with high efficiency, flexibility and portability. XGB has a number of unique features compared to other ensemble methods. It has an option to penalize complex models through both $L1$ and $L2$ regularization to prevent overfitting. A sparsity-aware split finding algorithm is incorporated into XGB to handle different types of sparsity patterns in the data. XGB has a distributed weighted quantile sketch algorithm to effectively handle weighted data. A block structure is embedded in its system to enable the data layout to be reused by subsequent iterations, instead of computing it again. Its out-of-core computing optimizes the available disk space and maximizes the usage of the computing when handling huge datasets that do not fit into memory. Although XGB appeared not long ago, it has become a widely used and very popular tool among Kaggle competitors and data scientists in both academic and industry \citep{Huang2018, Liu2019, Zhang2018}, as it has been extensively trialed for production on large-scale problems.

A similar hyperparameter optimization process was carried out for XGBoost. The optimal hyperparameters for mean pressure coefficients were set as follow. The minimum loss reduction required to make a further partition on a leaf node was set to 0. The learning rate, i.e. the step size shrinkage used in update to prevent overfitting, was set to 0.1. The maximum depth of a tree was set to 10. The sub-sample ratio of the training instances was set to 0.66. The best number of trees was set to 200. For the fluctuating pressure coefficients, the best value of the above hyperparameters were set to 0, 0.1, 12, 0.2, and 120 respectively. 

\subsection{Generative Adversarial Networks (GANs)}
\begin{figure*}[h!]
  \centering
  \includegraphics[width=\linewidth]{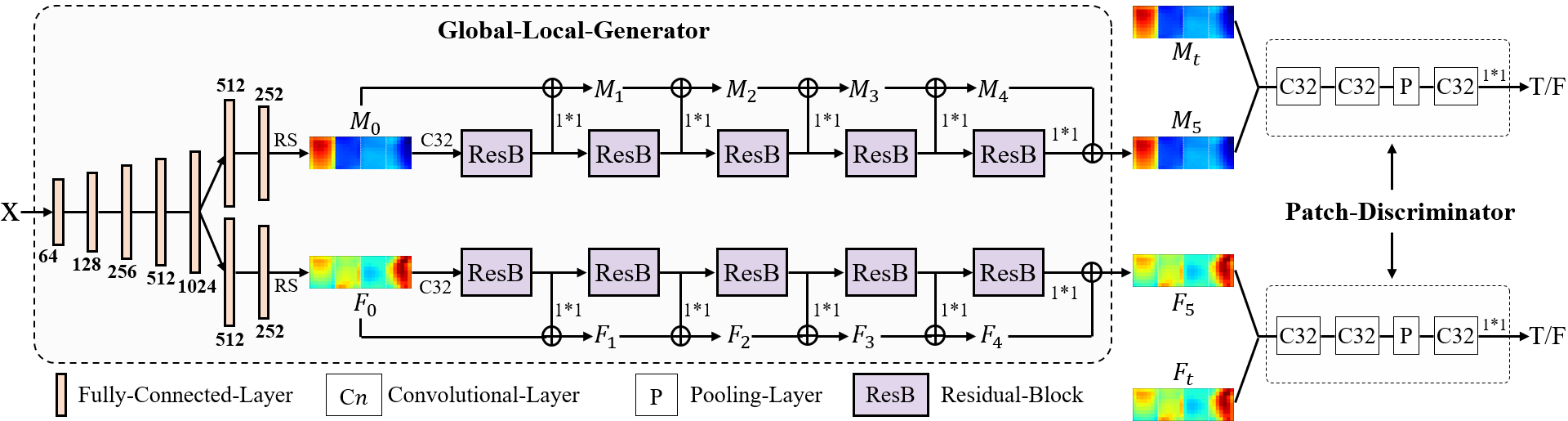}
  \caption{~Architecture of the tailor-designed two-stream GANs for wind pressure prediction. An orange pillar denotes a fully-connected layer. ``Cn'' denotes a convolutional layer with n filter and ``P'' is a max-pooling layer. $\oplus$ is an element-wise addition operation.}
  \label{fig5}
\end{figure*}

As one of the most promising newly-developed techniques in machine learning, generative adversarial networks (GANs) were proposed by \cite{Goodfellow2014} and have attracted worldwide interests for the task of generative modeling. In this section, a tailor-designed two-stream GANs was developed to simultaneously estimate both mean and fluctuating wind pressures on the principal building. As shown in Figure~\ref{fig5}, the GANs framework consists of a global-local generator and a patch discriminator. The generator was trained to produce the mean and fluctuating pressure maps that confuse the discriminator as much as possible. Meanwhile, the discriminator was trained to distinguish the ``fake'' maps generated by the generator from ``real'' maps. With this adversarial learning, the generative ability of the generators and the discriminative ability of the discriminators were gradually improved. Finally, high-quality wind pressure maps can be expected to be generated by the well-trained generator.

\begin{figure}[h!]
  \centering
  \includegraphics[width=0.75\linewidth]{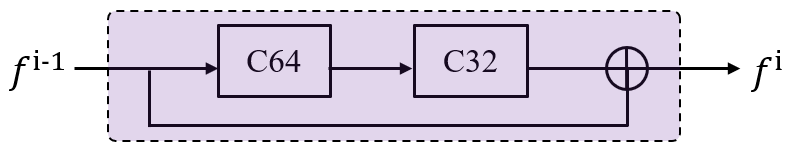}
  \caption{~Architecture of the Residual Block used in the generator. ``Cn'' denotes a convolutional layer with n filter. $\oplus$ is an element-wise addition operation.  $f^{i-1}$ and $f^i$ are the input feature and the output feature respectively. Both of them have 32 channels.}
  \label{fig6}
\end{figure}

\subsubsection{Global-Local Generator}
Given a condition $x = [S_x, S_y, \theta]$ where $(S_x, S_y)$ is the location of the interfering building and $\theta$ is the wind attack angle, the generator automatically generated the wind pressure maps on the principal building. Unlike previous methods that directly produce the expected images with a simple U-Net \citep{Ronneberger2015}, the pressure maps were generated in a global-local manner in this GANs.

First, the generator produces an initial mean pressure map $M_0$ and a fluctuating pressure map $F_0$ in a global generation manner. As shown in Figure~\ref{fig5}, the generator encodes the condition $x$ into a high-dimensional feature by feeding it through 5 FC layers. Specifically, the number of neural cells of these FC layers are 64, 128, 256, 512 and 1024 respectively. This high-dimensional feature was then fed into two branch networks, each of which consists of two FC layers (512 and 252 neural cells). The output of the first branch was reshaped to generate $M_0 \in R9*28$ and the output of the second branch was reshaped to obtain $F_0 \in R9*28$. 

Second, the quality of the generated mean pressure maps and fluctuating pressure maps was improved in a local refinement manner. A standard 3*3 convolutional layer with 32 filters was first utilized to encode $M_0$ and the output feature was denoted as $f^0 \in R32*9*28$. To learn the spatial local correlation on the pressure maps, $f^0$ was then fed into five Residual Blocks \citep{He2016}. As shown in Figure~\ref{fig6}, the Residual Block consisted of two convolutional layers and its output was computed by summing the input and the response map of the second convolutional layer. For convenience, the output feature of the $i^{th}$ Residual Block was denoted as $f^i \in R32*9*28$. Following \cite{Xie2015}, a side output was produced after each Residual Block. Specifically, the side-output mean pressure map $M_i$ of the $i^{th}$ Residual Block was computed as:
\begin{eqnarray}
M_i=w*f_i+M_{i-1}
\end{eqnarray}
where * denoted a convolution operation and $w$ was the parameter of a 1*1 convolutional layer with 1 filter. $M_5$ was the final mean pressure map. Meanwhile, $F_0$ was refined by another five Residual Blocks and the final fluctuating pressure map $F_5$ was obtained. 

\subsubsection{Patch Discriminator}
It is well known that the widely-used $L1$ loss and $L2$ loss would produce blurry results in image generation \citep{Larsen2016}. To alleviate the blurry effect on the generate wind pressure maps, their qualities were improved with a discriminator, which penalized the structure of the generated maps and can be understood as a form texture/style loss \citep{Larsen2016}. The discriminator was used to distinguish the generated maps from real maps. With adversarial learning, the generator was forced to produce pressure maps with high frequencies, in order to confuse the discriminator as well as possible.

In the present study, a Patch Discriminator, which aims to classify the local patches in the given wind pressure map as real or fake, was developed as shown in Figure~\ref{fig5}. Specifically, the Patch Discriminator consists of three 3*3 convolutional layers with 32 filters. Before the third convolutional layer, a 3*4 max-pooling layer was inserted to reduce the resolution of the feature map. Finally, a 1*1 convolutional layer was used to predict the probability of each local patch as a real sample. Two Patch Discriminators were trained, one took the given 9*28 mean pressure map as input and other fed the given 9*28 fluctuating pressure map into network. The output of each discriminator was a 3*7 probability map, each pixel on which denotes the probability of a 3*4 patch as a real local pressures map. 

\subsubsection{Implementation Details}
The proposed GANs was implemented with PyTorch toolbox \citep{Paszke2017}. The convolutional layers and fully-connected layers were initialized by a Gaussian distribution with zero mean and standard deviation of 0.02. The relevant hyperparameters were optimized via the 10-folder CV method and set as followings. The batch size was set to 32. Adam optimization \citep{Kingma2014} was used to optimize the networks for 2000 epochs. The learning rate was set to 0.0001 for the first 1000 epochs and it was linearly decayed to zero during the remaining 1000 epochs.

Moreover, adversarial loss was used to train the proposed GANs. When optimizing the generator, the parameters of the discriminators were fixed and vice versa. For the Global-Local Generator, its training loss $L_G$ was defined as:
\begin{eqnarray}
L_G^e=\frac{1}{6}\sum\limits_{i=0}^{5}(||M_t-M_i||^2+||F_t-F_i||^2)
\end{eqnarray}
\begin{eqnarray}
L_G^a=log(D_M(M_5))+log(D_F(F_5))
\end{eqnarray}
\begin{eqnarray}
L_G=L_G^e+\alpha*L_G^a
\end{eqnarray}
where $L_G^e$ was the Euclidean loss between the generated pressure maps and their ground-truth $M_t$ or $F_t$, and $L_G^a$ was the adversarial discriminative loss. The generated map $M_5$ was fed into the discriminator of mean pressure coefficients and the output probability map was denoted as $D_M$ ($M_5$). $M_5$ was expected to cheat the discriminator, which means the discriminator treats the generated map as real sample, thus the loss was defined as $log (D_M(M_5))$. With the same principle, the discriminative loss of fluctuating pressure coefficients was defined as $log (D_F(F_5))$. $\alpha$ was used to adapt the weight coefficient of the adversarial discriminative loss and it was set to 100 in the present study. For the Patch Discriminator, its training loss $L_D$ was defined as:
\begin{eqnarray}
L_D^M=log(D_M(M_t))+log(1-D_M(M_5))
\end{eqnarray}
\begin{eqnarray}
L_D^F=log(D_F(F_t))+log(1-D_F(F_5))
\end{eqnarray}
\begin{eqnarray}
L_D=L_D^M+L_D^F
\end{eqnarray}
where $L_D^M$ and $L_D^F$ were the common cross entropy loss. This discriminative loss $L_D$ forced the discriminator to distinguish the generated maps from real maps as well as possible.

\section{Results and analyses }\label{sec4}
\subsection{Surface pressure prediction on principal building}
\begin{figure*}
     \centering
     \begin{subfigure}[b]{0.3\linewidth}
         \centering
         \includegraphics[width=\textwidth]{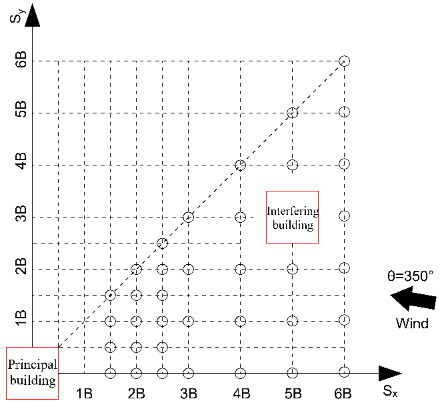}
         \caption{Interference configuration.}
         \label{fig7a}
     \end{subfigure}
     \hfill
     \begin{subfigure}[b]{0.3\linewidth}
         \centering
         \includegraphics[width=\textwidth]{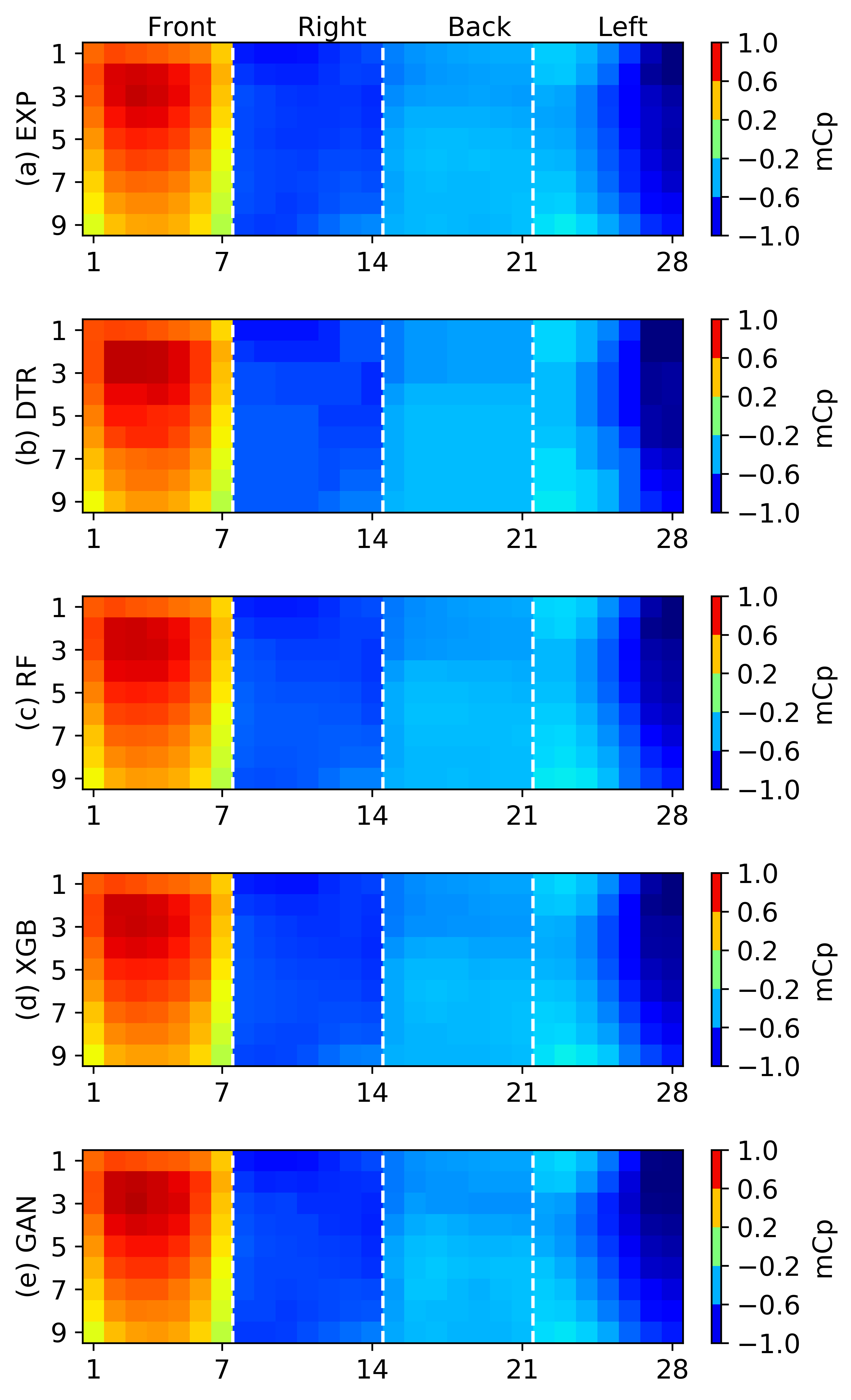}
         \caption{Mean pressure coefficients.}
         \label{fig7b}
     \end{subfigure}
     \begin{subfigure}[b]{0.3\linewidth}
         \centering
         \includegraphics[width=\textwidth]{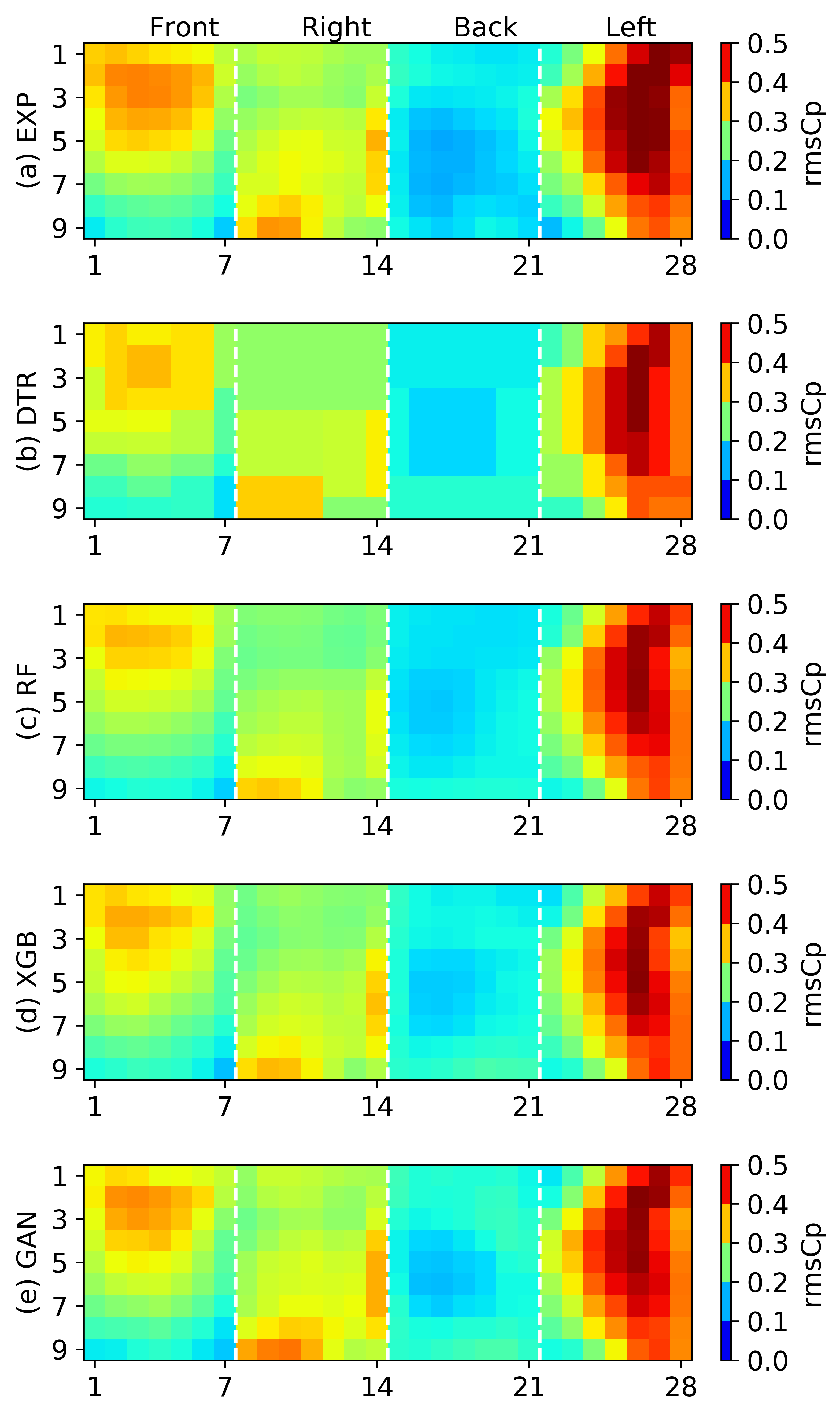}
         \caption{Fluctuating pressure coefficients.}
         \label{fig7c}
     \end{subfigure}
     \hfill
     \caption{~Comparison between pressure distributions on four faces of principal building predicted by machine learning models and experimental data with interfering building located at $S_x = 5.0B, S_y = 3.0B$ and under 350$^\circ$ wind attack angle.}
     \label{fig7}
\end{figure*}

\begin{figure*}
     \centering
     \begin{subfigure}[b]{0.3\linewidth}
         \centering
         \includegraphics[width=\textwidth]{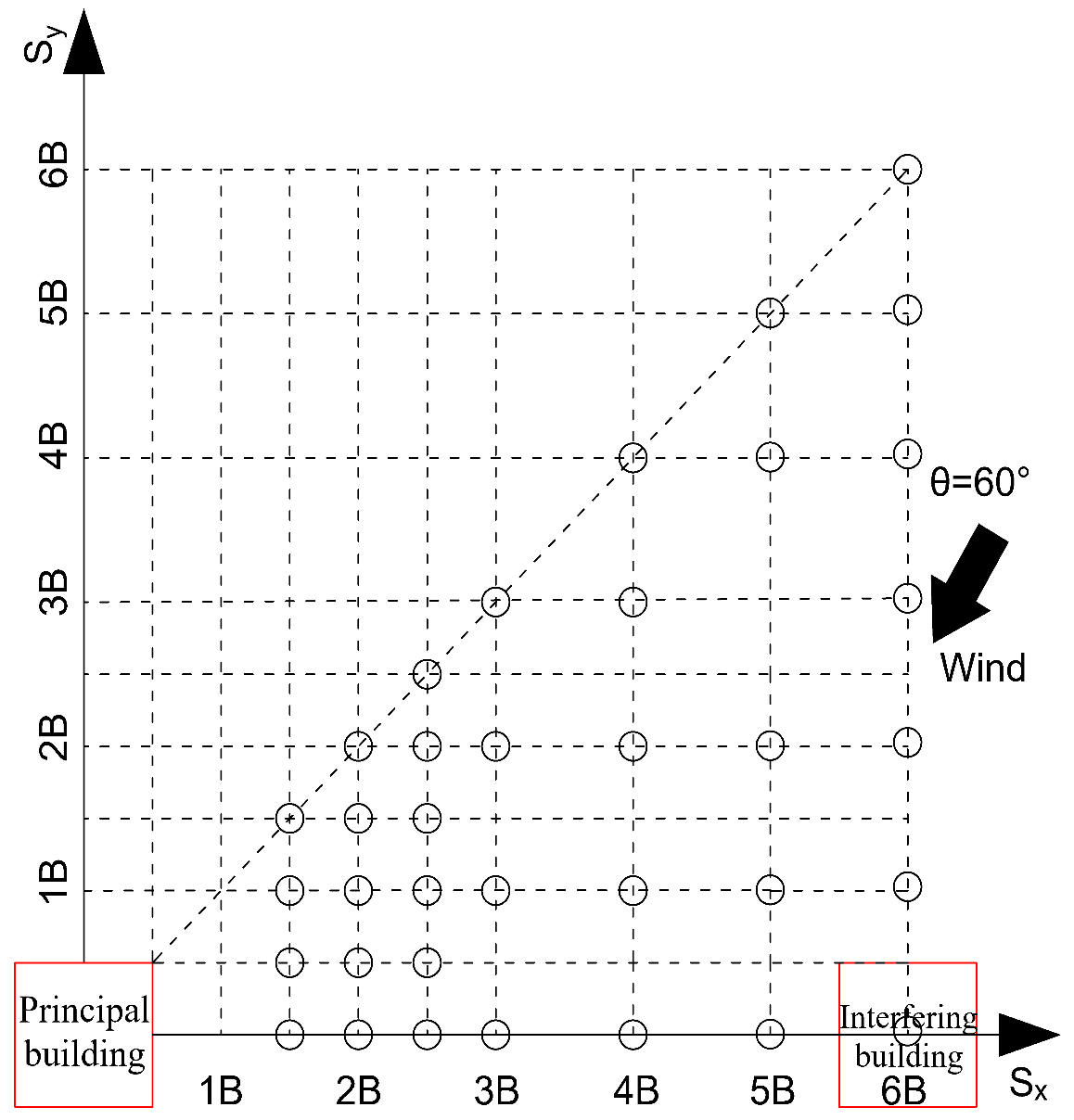}
         \caption{Interference configuration.}
         \label{fig8a}
     \end{subfigure}
     \hfill
     \begin{subfigure}[b]{0.3\linewidth}
         \centering
         \includegraphics[width=\textwidth]{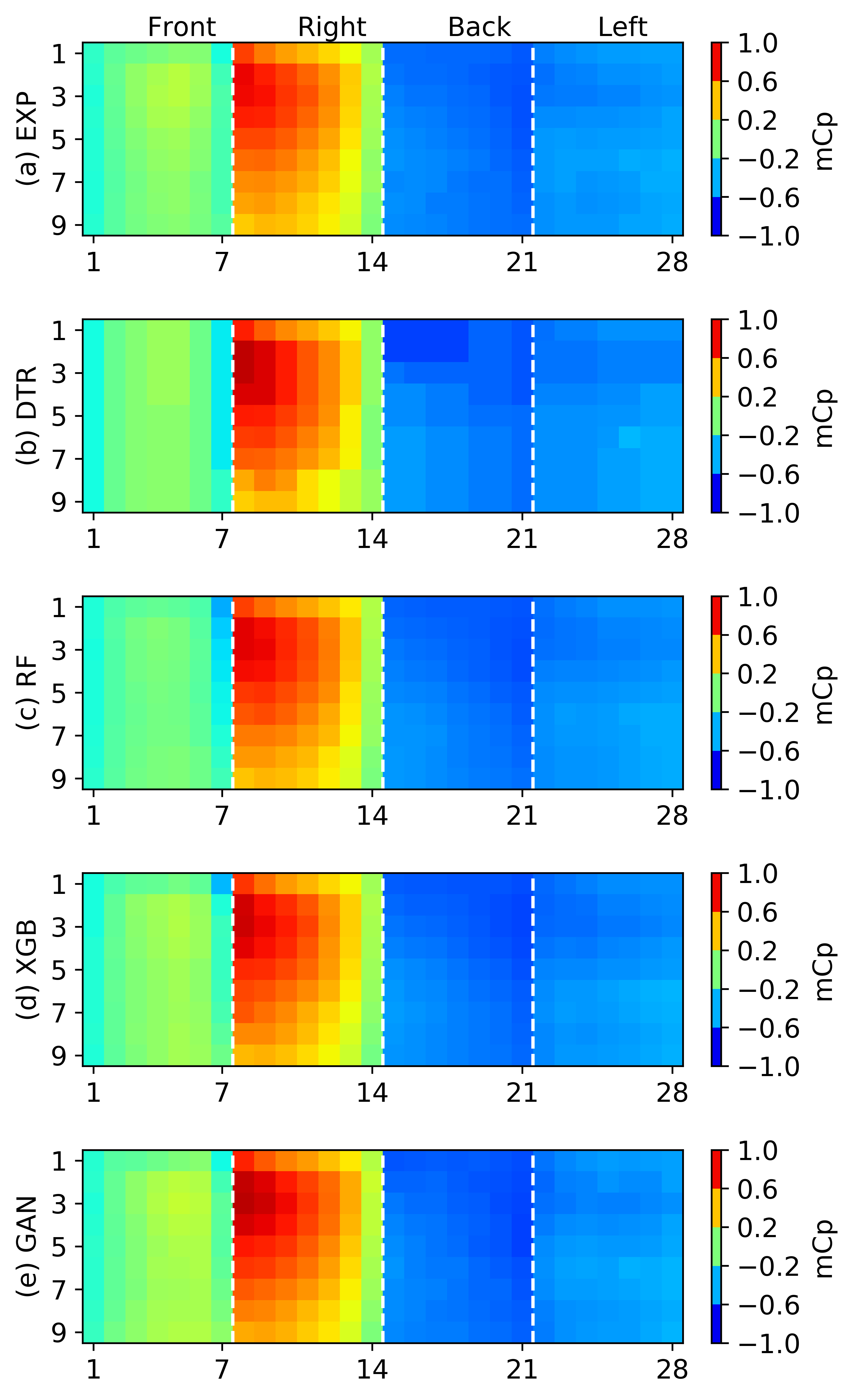}
         \caption{Mean pressure coefficients.}
         \label{fig8b}
     \end{subfigure}
     \hfill
     \begin{subfigure}[b]{0.3\linewidth}
         \centering
         \includegraphics[width=\textwidth]{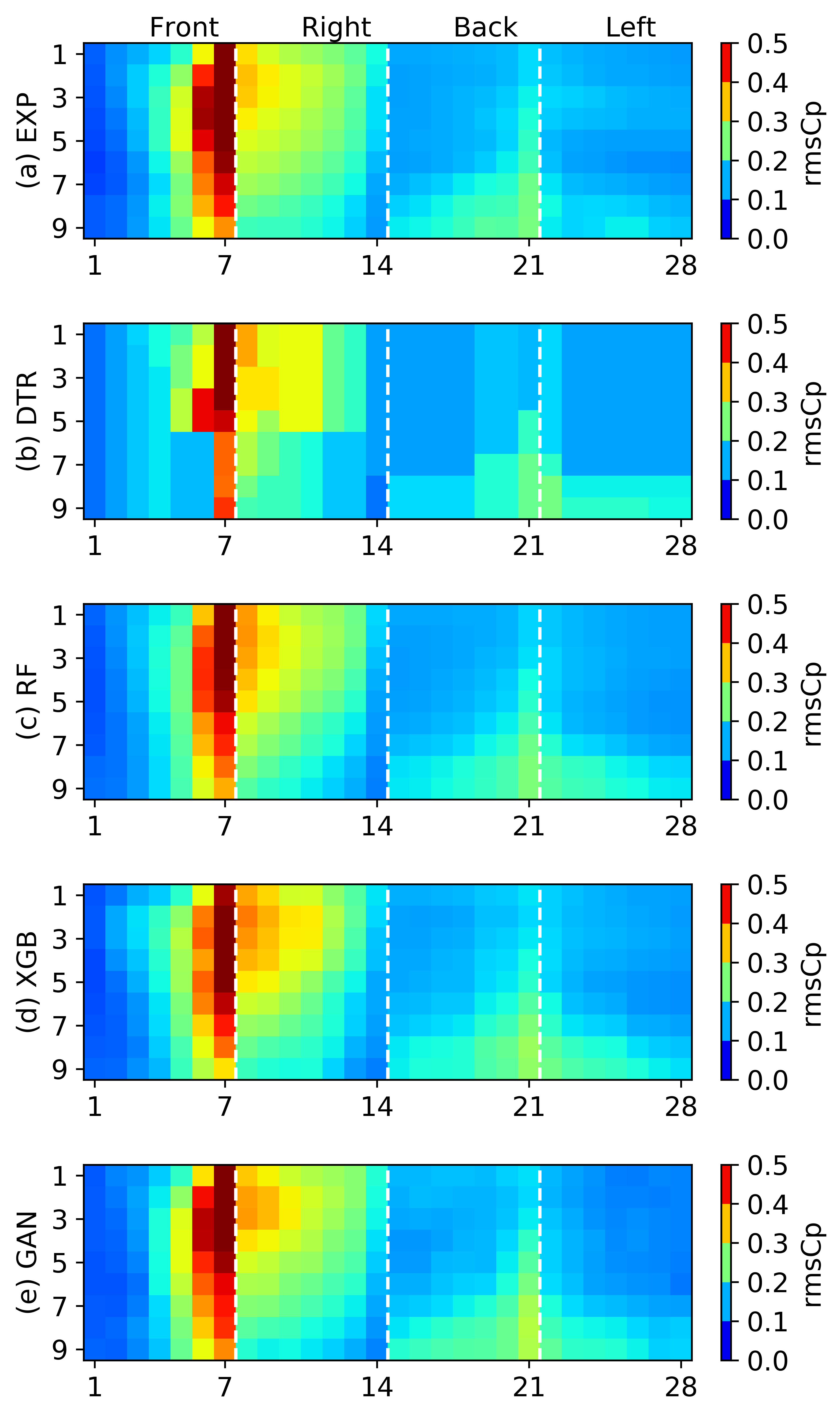}
         \caption{Fluctuating pressure coefficients.}
         \label{fig8c}
     \end{subfigure}
     \caption{~Comparison between pressure distributions on four faces of principal building predicted by machine learning models and experimental data with interfering building located at $S_x = 6.0B, S_y = 0B$ and under 60$^\circ$ wind attack angle.}
     \label{fig8}
\end{figure*}

To compare the performances of the ML models, pressure distributions on the principal building of the 6 cases set aside before training the models were predicted by the models and compared with wind tunnel experimental results. As mentioned above, these 6 cases were randomly selected at the very beginning of this ML study and set aside for this final validation. That is to say, these models have never seen the pressure data of these 6 cases before. Results of only two representative cases of the 6 cases are shown in Figure~\ref{fig7} and Figure~\ref{fig8} due to page limit. For the case in Figure~\ref{fig7}, the interfering building is located at $S_x = 5.0B, S_y = 3.0B$, and the wind attack angle is 350$^\circ$; for the case in Figure~\ref{fig8}, the location of the interfering building is $S_x = 6.0B, S_y = 0$, and the attack angle is 60$^\circ$. 

In Figure~\ref{fig7}(b), as expected, mean pressure coefficients on the front face are positive since wind strikes the front face directly. The stagnation region slightly deviates to left from middle due to the 10$^\circ$ deviation of the wind direction from the front-back direction. It can be seen that the four models all reproduced this deviation. The other three faces are immersed in the flow separation region and wake region, and hence exhibit negative pressures. The wind attack angle of 350$^\circ$ creates asymmetrical pressure distributions on the two side faces, which is also captured by all the four models. In general, the four models accurately predict mean pressure coefficients over the most region of the four faces. 

In Figure~\ref{fig7}(c), the fluctuating pressure coefficients on the left face are particularly large, which results from the recirculation bubble created by the shear layer separated from the leading edge. On the stagnation region of the front face, the coefficients are slightly larger than the adjacent region due to the oncoming flow turbulence. Overall, these features are roughly reproduced by the four models. In contrast, the prediction of the GANs model is much closer to the experimental data than the other three models. Similarly, the GANs model exhibits a higher accurate prediction on the other two faces. On the other hand, the discrepancies between the predictions and the experimental results are more obvious than those in the mean pressure coefficients. Despite visible discrepancies, the ML models generally reproduce these fluctuating pressure coefficients. 

Figure~\ref{fig8} presents comparisons between predictions of pressure coefficients using the models and experimental results of the case with the location of the interfering building of $S_x = 6.0B, S_y = 0B$, and the attack angle of 60$^\circ$. The stagnation region appears on the right face and is close to the leading edge (see Figure~\ref{fig8}(b)). The flow separates from this edge and leads to a recirculation bubble on the front face close to this edge. The recirculation bubble induces strong pressure fluctuation in this region as shown in Figure~\ref{fig8}(c). The four models generally capture all these features. For mean pressure coefficients, the predictions by the four models are excellent and match very well with the experimental data. In contrast, the predictions of the fluctuating pressure coefficients are not as good as those of the mean pressure coefficients. The discrepancies are visible, in particular the predictions of the DTR model. For example, the DTR model underestimates the fluctuating pressure coefficients on the aforementioned recirculation zone of the front face. In contrast, the GANs model predicts these coefficients very well. 

\subsection{Performance comparisons of machine learning models}
\begin{figure}
     \centering
     \begin{subfigure}{1\linewidth}
         \centering
         \includegraphics[width=\textwidth]{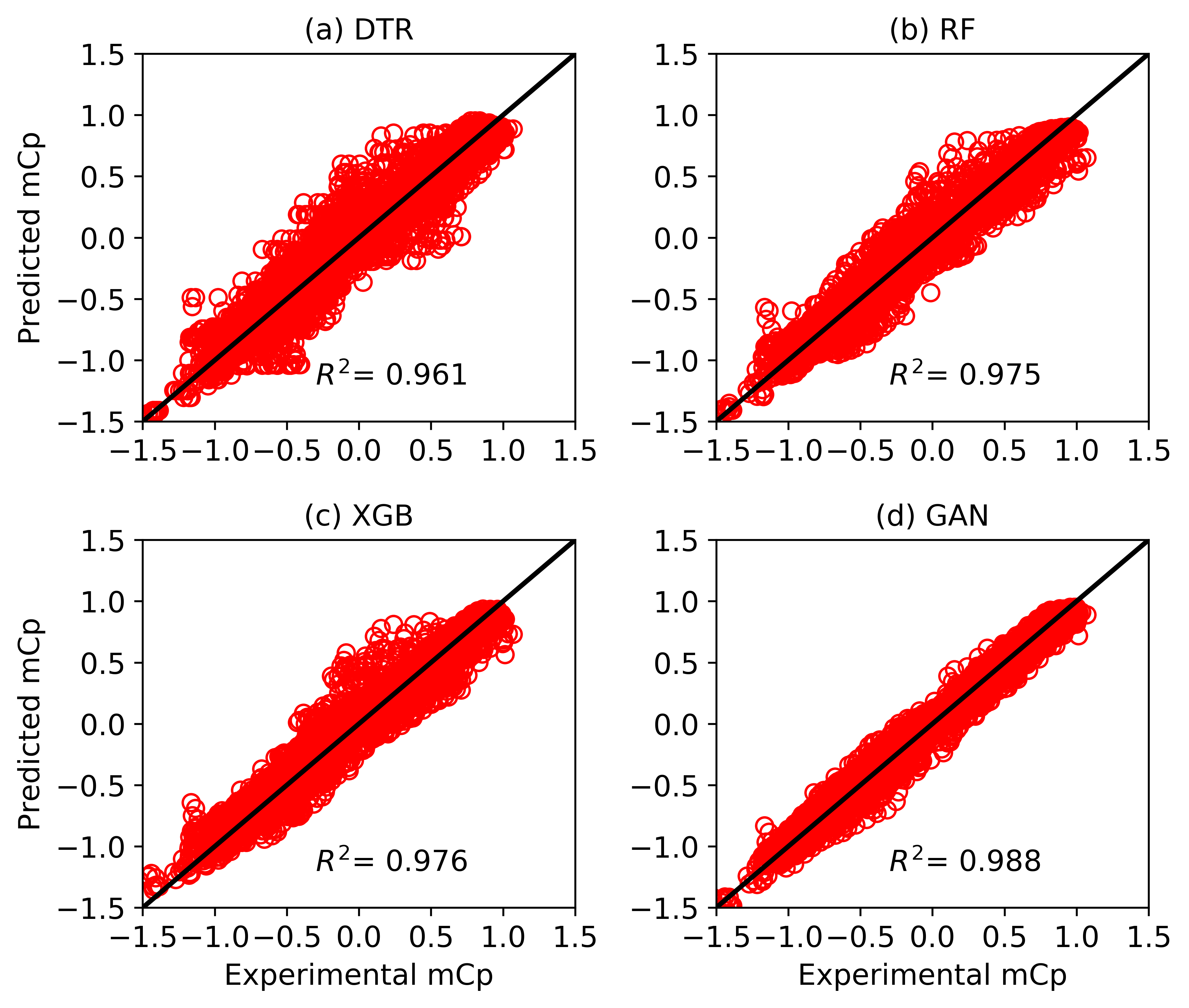}
         \caption{Mean pressure coefficient (mCp)}
         \label{fig9a}
     \end{subfigure}
     \hfill
     \begin{subfigure}{1\linewidth}
         \centering
         \includegraphics[width=\textwidth]{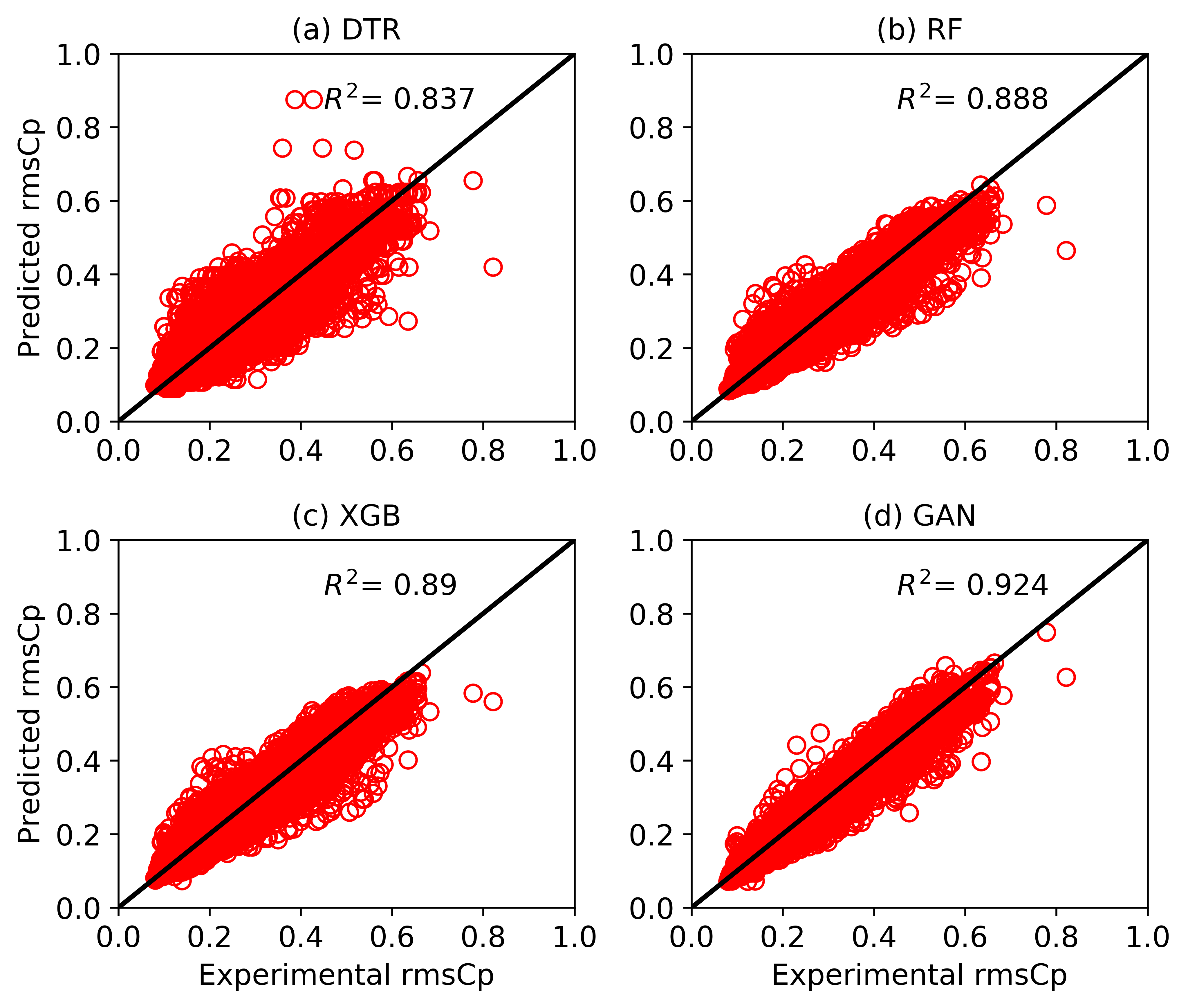}
         \caption{Fluctuating pressure coefficient (rmsCp)}
         \label{fig9b}
     \end{subfigure}
     \caption{~Comparisons between pressure coefficients of testing dataset and pressure coefficients predicted by three machine learning models.}
     \label{fig9}
\end{figure}

The above observations have preliminarily demonstrated the capability of the four models in predicting both mean and fluctuating pressure coefficients on the principal building under 2 unseen interference conditions. To further evaluate the performances of the four models in detail, the four models were trained based on the entire training dataset (i.e. 80\% of the 30\% dataset) by using the optimal hyperparameters. The models were then validated by the testing dataset, i.e. 20\% of the 30\% dataset. Comparisons between pressure coefficients predicted by the four models and the testing dataset are shown in Figure~\ref{fig9}. It can be seen that for mean pressure coefficients, all four models exhibit very high $R^2$ scores, higher than 0.96. The DTR model shows the lowest score. The two ensemble models, i.e. RF and XGB, exhibit higher scores than the DTR model as expected. The GANs model wins the highest score, $R^2$=0.988, much higher than the other three models. 

\begin{figure}
     \centering
     \begin{subfigure}{0.8\linewidth}
         \centering
         \includegraphics[width=\textwidth]{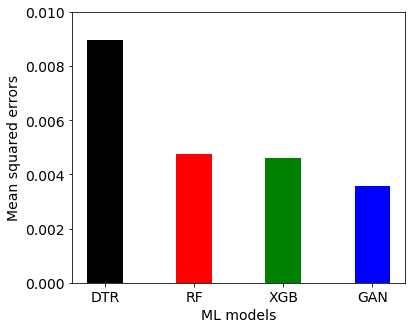}
         \caption{Mean pressure coefficient (mCp)}
         \label{fig10a}
     \end{subfigure}
     \hfill
     \begin{subfigure}{0.8\linewidth}
         \centering
         \includegraphics[width=\textwidth]{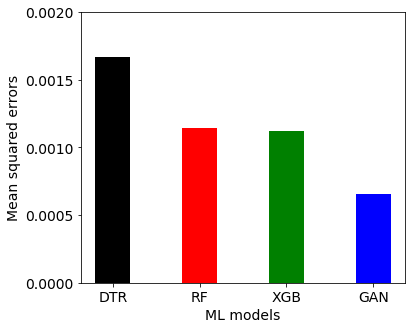}
         \caption{Fluctuating pressure coefficient (rmsCp)}
         \label{fig10b}
     \end{subfigure}
     \caption{~Comparison of mean square errors of ML models for predicting pressure coefficients.}
     \label{fig10}
\end{figure}

In contrast, the $R^2$ scores of the four models for fluctuating pressure coefficients are lower than those for the mean pressure coefficients. Likewise, the GANs model shows the highest score, $R^2$=0.924, and is much better than other three models of which the scores are all lower than 0.9. Another metric of evaluating the model performance, mean squared error, was also used in this study as shown in Figure~\ref{fig10}. The superior performance of the GANs model in both mean and fluctuating pressure coefficients is obvious. 

It is not surprising that the GANs models exhibit the best performance for both mean and fluctuating coefficients. GANs, well known with strong generative ability based on limited samples, have been proven quite effective in computer vision and image processing \citep{Ledig_2017_CVPR}. In the present study, the output of the GANs model was an image representing the pressures on all faces of the building, while the output of the other ML models was the pressure at one measured point on the surface of the building. This fact endowed the GANs model with the ability of considering the spatial correlation of the pressure on different faces of the building. Taking the spatial correlation of the pressure on different faces into account surely augmented the performance of the ML model. Furthermore, the above substantial efforts of adopting various strategies to build the tailor-designed GANs improved the performance of the GANs model in predicting both mean and fluctuating pressure coefficients. Thus, the GANs model with its optimal hyperparameters is selected as the best model to predict both mean and fluctuating pressure coefficients on the principal building in the following section. 

\subsection{Effect of different portions of data used on performance of machine learning model }
\begin{figure}[h!]
  \centering
  \includegraphics[width=\linewidth]{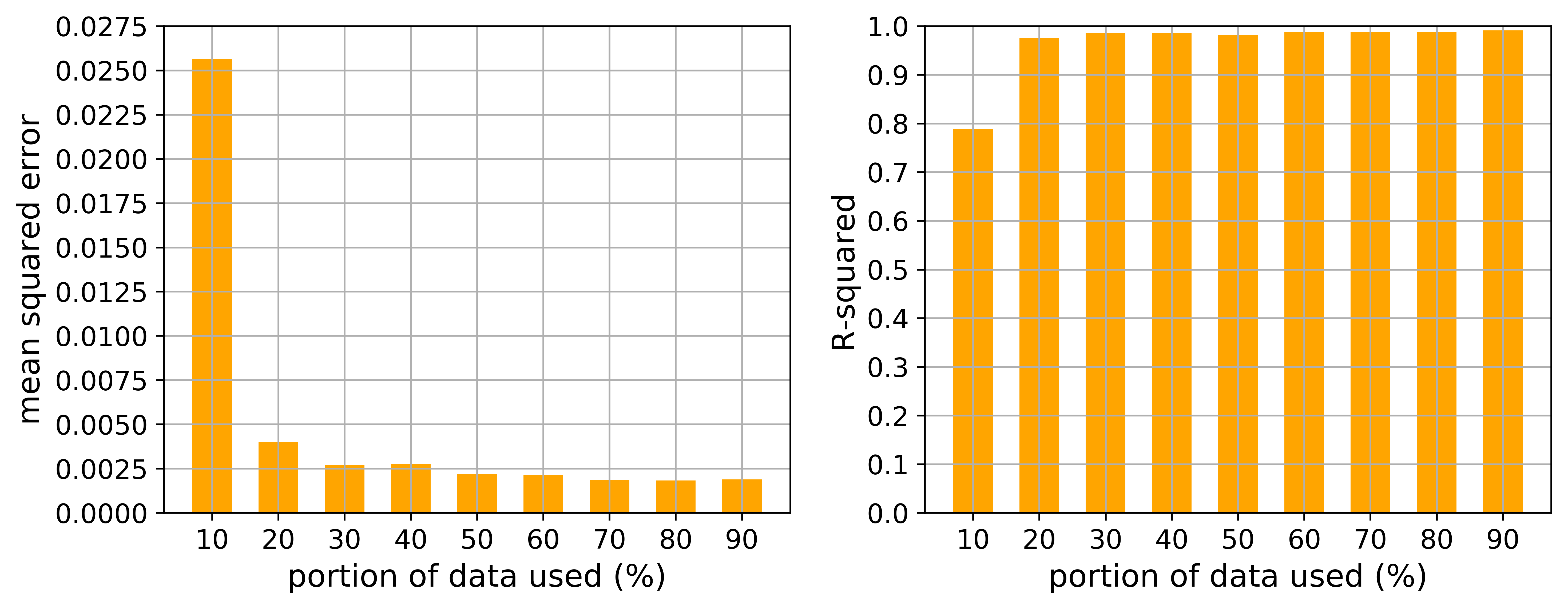}
  \caption{~Mean squared error and R-squared against portion of data used in training GANs for mean pressure coefficients.}
  \label{fig11}
\end{figure}

\begin{figure}[h!]
  \centering
  \includegraphics[width=\linewidth]{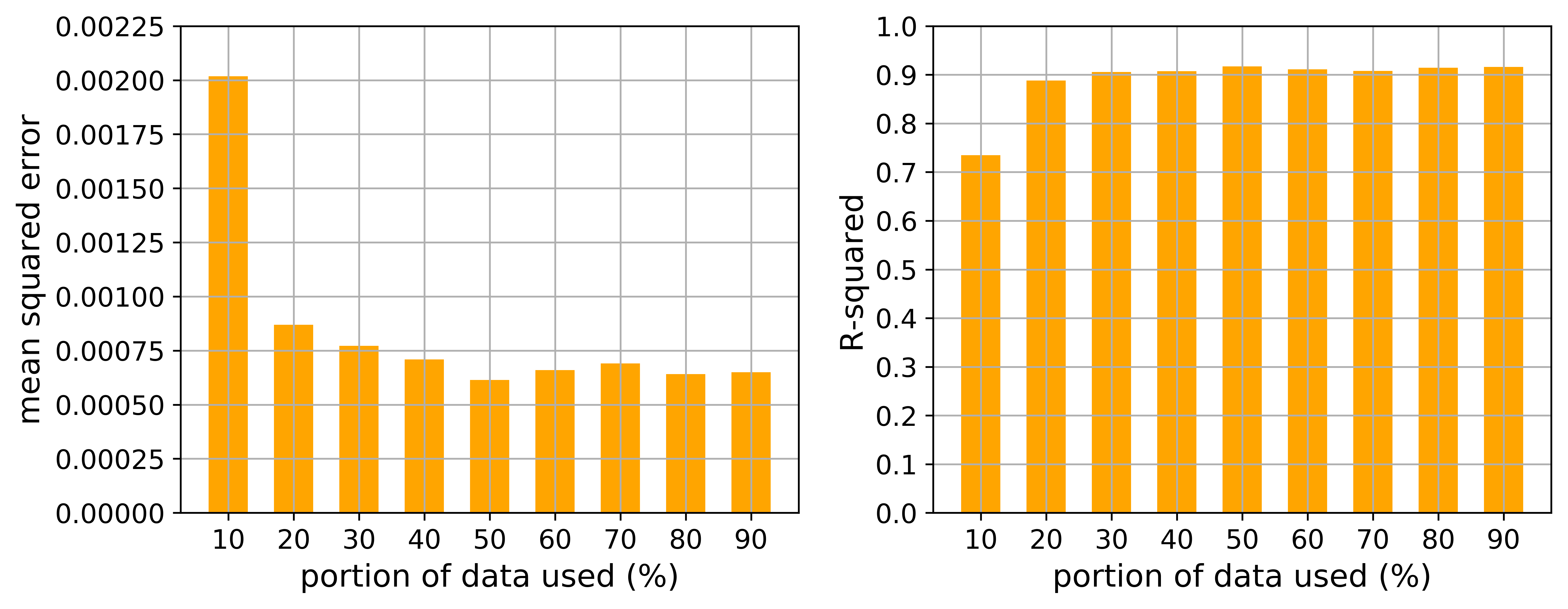}
  \caption{~Mean squared error and R-squared against portion of data used in training GANs for fluctuating pressure coefficients.}
  \label{fig12}
\end{figure}

So far, the above GANs model is trained based on only 30\% of 2558 cases, i.e. the entire wind tunnel test cases excluding 6 cases set aside for validation. As suggested in the Introduction, this study aims to reveal the interference effects on buildings under various interference conditions including untested conditions based on limited wind tunnel test dataset by using ML techniques. In order to explore how much data is sufficient to build an accurate GANs model, effects of different portions of data used in training on the model performance are evaluated in this section. The tested portions range from 10\% to 90\% in a 10\% interval. The TTV process is also adopted in training and testing the model. 80\% of the selected portion of the dataset are used to train the ML model and the rest are used to test the model. The validation results including mean squared errors and R-squared values by using the 20\% of the selected portion are shown in Figure~\ref{fig11} and Figure~\ref{fig12} for mean and fluctuating pressure coefficients respectively. 

As can be seen in Figure~\ref{fig11}, the mean squared error for the portion of 10\% is much higher than those for other portions. Increasing the portion to 20\% dramatically reduces the mean squared error. Further increasing the portion to 30\%, an obvious reduction is still observed. In contrast, although an increase beyond 30\% in the portion still reduces the mean squared error, the reduction is not very significant. Similarly, the R-squared value exhibits an abrupt rise after increasing the portion from 10\% to 20\%. A slight increase can still be observed from 20\% to 30\%. A further increase in the R-squared value by increasing the portion is no longer remarkable. 

The effect of the portion of data used in training the GANs model on its performance for fluctuating pressure coefficients is very similar to the effect for mean pressure coefficients. As shown in Figure~\ref{fig12}, the model training based on only 10\% of the dataset leads to a very large mean squared error and very low R-squared value. Beyond 20\%, although these two metrics change with increasing the portion, the change is not remarkable. 

\begin{figure}[h!]
  \centering
  \includegraphics[width=\linewidth]{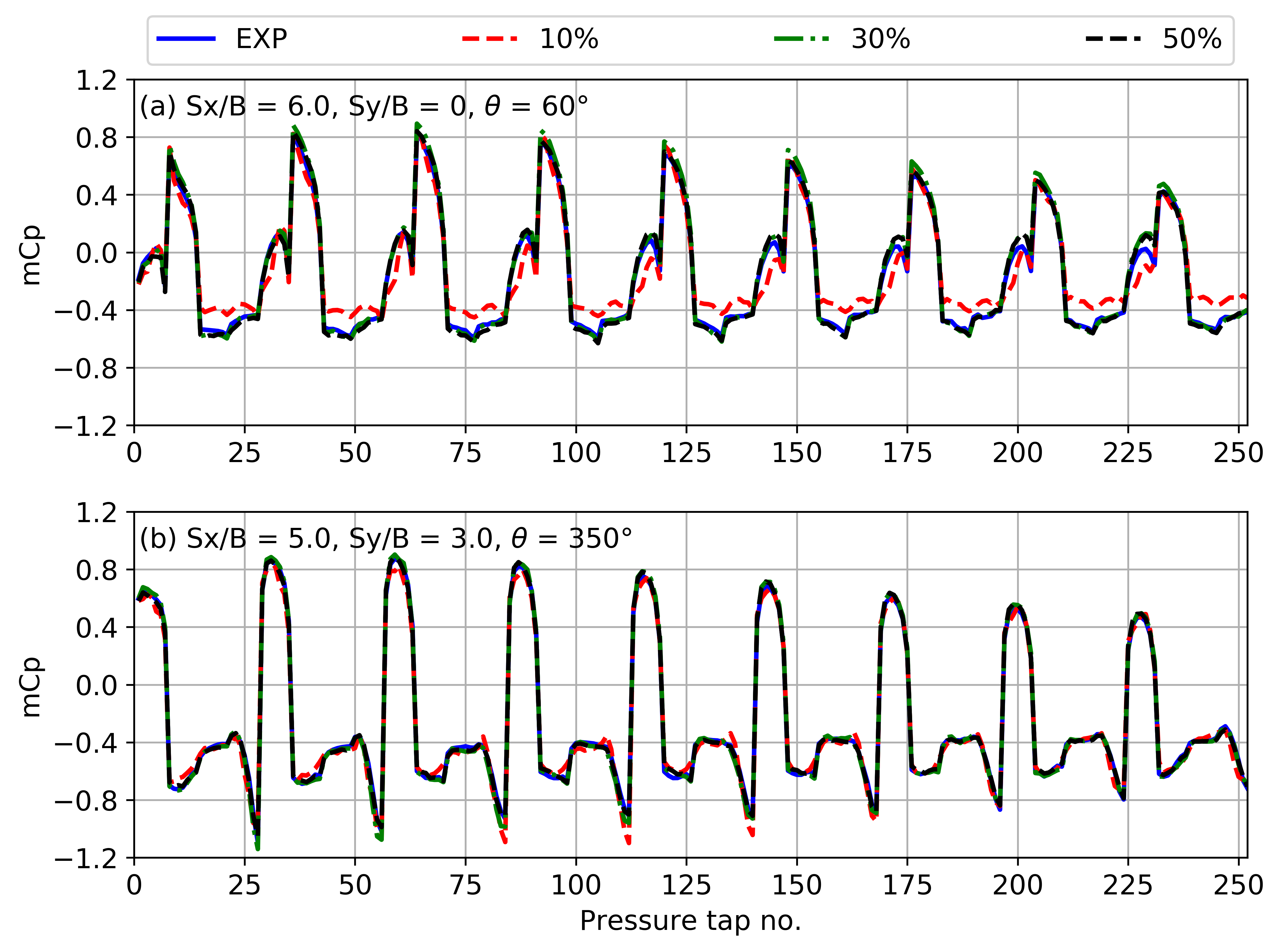}
  \caption{~Comparisons of mean pressure coefficients between experimental data and GANs predictions using different portion of data.}
  \label{fig13}
\end{figure}

\begin{figure}[h!]
  \centering
  \includegraphics[width=\linewidth]{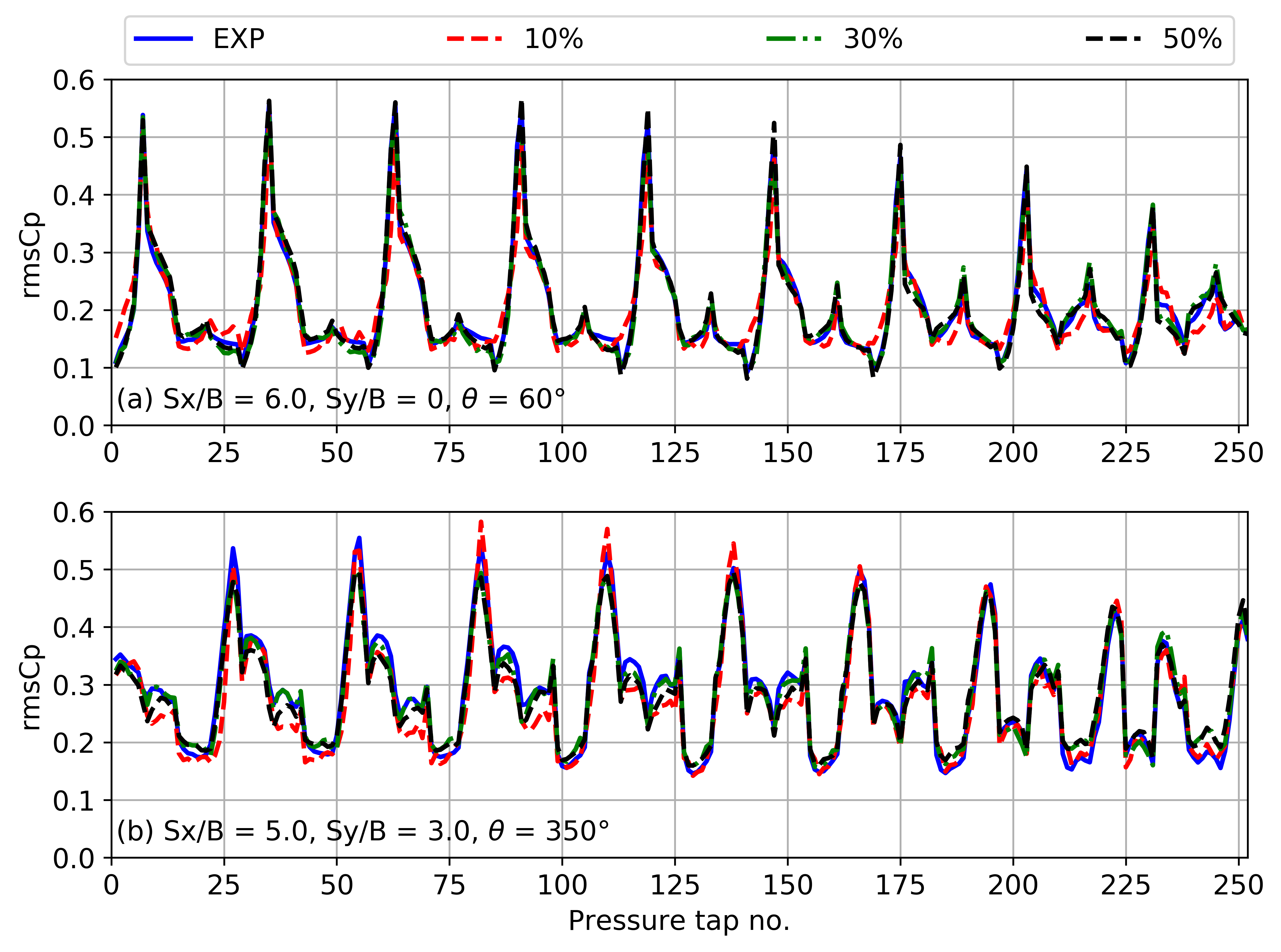}
  \caption{~Comparisons of fluctuating pressure coefficients between experimental data and GANs predictions using different portion of data.}
  \label{fig14}
\end{figure}

To further illustrate the effect of the portion of the dataset on the performance of the GANs model, three GANs models trained based on 10\%, 30\%, 50\% of the dataset are used to predict mean and fluctuating pressure coefficients of the 6 cases set aside for final validation. Comparisons between experimental data and predictions made by the three GANs models are presented for 2 out of the 6 cases, as shown in Figures~\ref{fig13} and ~\ref{fig14}. As shown in Figure~\ref{fig13}(a), the predictions of the GANs model based on 10\% of the dataset deviate from the experimental data. In contrast, the predictions of the GANs models based on both 30\% and 50\% of the dataset agree very well with the experimental data. Similarly, in Figure~\ref{fig13}(b), visible discrepancies between predictions made by the GANs model based on 10\% of the dataset and the experimental data are observed. However, the other two GANs models predict the mean pressure coefficient very accurately. Furthermore, the predictions of the GANs model based on 30\% of the dataset are as accurate as the GANs model based on 50\% of the dataset. 

Comparisons of fluctuating pressure coefficients between experimental data and the GANs predictions are shown in Figure~\ref{fig14}. In general, the predictions are not as accurate as those of mean pressure coefficients as shown in Figure~\ref{fig13}, which agrees with the statistical observation in section 4.1. Similar to the observations in Figure~\ref{fig13}, the GANs model based on 10\% of the dataset is less accurate than the other two GANs models. The deviations of the predictions made by this GANs model is more obvious. The other two models exhibit an equal performance.  

Overall, more data fed into the GANs model lead to a stronger model which can predict both mean and fluctuating pressure coefficients more accurately. This performance improvement is very prominent when the portion is less than 30\%. However, when the portion exceeds 30\%, the performance improvement is much less remarkable. Nevertheless, the GANs model based on 30\% of the dataset is capable of predicting both mean and fluctuating pressure coefficients quite accurately. 

\section{Discussions}\label{sec5}
To fully understand the interference effects of buildings, it often requires a large amount of wind tunnel tests, which is not always feasible due to high cost of experimental tests. However, limited wind tunnel tests are unable to fully reveal the interference effects. To resolve this contradiction, ML techniques have been used to fully reveal the interference effects on buildings under various interference conditions including untested conditions based on limited wind tunnel test dataset. The test results indicate that the GANs model trained based on only 30\% of the entire 2664 dataset from the TPU aerodynamic database is capable of accurately predicting wind pressures on the principal building under unseen interference conditions. That is to say, the GANs model based on 30\% of the dataset can be considered equivalent to the entire 2664 dataset. By using this GANs technique, 70\% of the wind tunnel test cases can be saved, which largely alleviates the cost of this kind of study.

The above validation processes have proven the capability of the GANs model in predicting wind pressures on the principal building under any unseen interference effects. Therefore, it is anticipated that based on this GANs model, high-resolution interference factor maps of various force and moment coefficients can be built, which is unlikely to be achieved based on wind tunnel tests alone.

\section{Conclusions}\label{sec6}
This study has employed four machine learning algorithms, i.e. decision tree, random forest, XGBoost, and generative adversarial networks (GANs), in evaluating interference effects of buildings. The dataset used in this study is obtained from the aerodynamic database of Tokyo Polytechnic University. The performances of the four models built using the above four algorithms are evaluated based on 30\% of the whole dataset. It was found that the GANs model exhibited the best performance in predicting both mean and fluctuating pressure coefficients. A number of GANs models are then trained based on various portions of the dataset ranging from 10\% to 90\% in a 10\% interval. In general, more data fed into the GANs model lead to a stronger model that can predict both mean and fluctuating pressure coefficients more accurately. This performance improvement is very remarkable when the portion is less than 30\%. However, beyond 30\%, the improvement is much less prominent and even negligible. A validation process has proven that the GANs model based on 30\% of the dataset is capable of predicting both mean and fluctuating pressure coefficients quite accurately. More importantly, the GANs model can quickly and accurately predict the pressure coefficients on the principle building under various unseen interference conditions. That is to say, the GANs model based on 30\% of the dataset is superior to the full dataset with the entire 2664 testing cases. Therefore, by using this ML technique, 70\% of the wind tunnel test cases can be saved, which largely alleviates the cost of this kind of wind tunnel testing study. Consequently, this GANs model can be used to resolve the conflicting requirement between limited wind tunnel tests that produce unreliable results and a completed investigation of the interference effects that is costly and time-consuming. In addition, it is anticipated that this GANs model can be used to generate high-resolution interference factor maps of various force and moment coefficients. 

\section*{Acknowledgments}
The authors gratefully acknowledge the aerodynamic database of Tokyo Polytechnic University. Without this database, the work is unlikely accomplished. The authors also acknowledge two open source machine learning and deep learning platforms, i.e. scikit-learn and PyTorch. A fruitful discussion with Dr. Wonsul Kim is greatly appreciated.  

\balance
\nocite{*}% Show all bib entries - both cited and uncited; comment this line to view only cited bib entries;
\bibliography{wileyNJD-APA}%

\end{document}